\documentclass{article}
\usepackage{spconf,amsmath}

\usepackage{multirow}
\usepackage{graphicx}
\usepackage{amsmath}
\usepackage[psamsfonts]{amssymb}
\usepackage{amsxtra}
\usepackage{threeparttable}
\usepackage{amsfonts}
\usepackage{mathrsfs}
\usepackage{arydshln}
\usepackage{cite}
\usepackage{enumitem}
\usepackage{url}

\makeatletter
\let\MYcaption\@makecaption
\makeatother
\usepackage{subcaption}
\makeatletter
\let\@makecaption\MYcaption
\makeatother
\makeatletter
\renewcommand\section{\@startsection{section}{1}{\z@}
                      {0.5ex \@plus 0ex \@minus -2ex}
                      {0.5ex \@plus 0ex}
                      {\normalfont\Large\bfseries}}
\renewcommand\subsection{\@startsection{subsection}{2}{\z@}
                      {0.5ex \@plus 0ex \@minus -2ex}
                      {0.5ex \@plus 0ex}
                      {\normalfont\large\bfseries}}
\renewcommand\subsubsection{\@startsection{subsubsection}{3}{\z@}
                      {0.5ex \@plus 0ex \@minus -2ex}
                      {0.5ex \@plus 0ex}
                      {\normalfont\normalsize\bfseries}}
\def\@listi{\leftmargin\leftmargini
            \parsep 1.0pt
            \topsep 0.2\baselineskip \@minus 0.1\baselineskip
            \itemsep 1.0pt \relax}
\let\@listI\@listi
\makeatother

\newcommand{\myset}[1]{\mathrm{#1}}

\newcounter{num}
\newcommand{\rnum}[1]{\setcounter{num}{#1} \roman{num}}

\title{Image Enhancement Network Trained by Using HDR images}
\name{Yuma Kinoshita and Hitoshi Kiya}
\address{Tokyo Metropolitan University, Tokyo, Japan}
\begin{document}\sloppy
\setlength{\parskip}{0.0pt}
\setlength{\tabcolsep}{1.0pt}
\setlength{\floatsep}{0.0pt}
\setlength{\abovecaptionskip}{0.0pt}
\setlength{\intextsep}{0.0pt}
\setlength{\dblfloatsep}{0.0pt}
\setlength{\dbltextfloatsep}{1.0pt}
\abovedisplayskip=0pt
\belowdisplayskip=0pt
%
\maketitle
\begin{abstract}
  In this paper, a novel image enhancement network is proposed,
  where HDR images are used for generating training data for our network.
  Most of conventional image enhancement methods,
  including Retinex based methods,
  do not take into account restoring lost pixel values caused by clipping and quantizing.
  In addition, recently proposed CNN based methods still have a limited scope of application
  or a limited performance, due to network architectures.
  In contrast, the proposed method have a higher performance and a simpler network architecture
  than existing CNN based methods.
  Moreover, the proposed method enables us to restore lost pixel values.
  Experimental results show that the proposed method can provides
  higher-quality images than conventional image enhancement methods
  including a CNN based method,
  in terms of TMQI and NIQE.
\end{abstract}
\renewcommand{\thefootnote}{\fnsymbol{footnote}}
\footnote[0]{This work was supported by JSPS KAKENHI Grant Number JP18J20326.}
\renewcommand{\thefootnote}{\arabic{footnote}}
\section{Introduction}
  The low dynamic range (LDR) of modern digital cameras
  is a major factor preventing cameras from capturing images as well as human vision.
  This is due to the limited dynamic range that imaging sensors have.
  This limit results in low contrast in images taken by digital cameras.
  The purpose of enhancing images is to show hidden details in such images.

  Various kinds of research on single-image enhancement have so far been reported
  \cite{zuiderveld1994contrast, wu2017contrast,
  guo2017lime, fu2016weighted, kinoshita2017pseudo, kinoshita2018pseudo}.
  Most of image enhancement methods can be divided into two methods:
  histogram equalization (HE) based methods and Retinex-based methods.
  However, both HE- and Retinex-based methods cannot restore lost pixel values due to
  quantizing and clipping.
  For this reason, a few image enhancement method based on
  convolutional neural networks (CNNs)
  have recently been developed \cite{chen2018learning, yang2018image}.

  CNN based methods can solve the problem that traditional methods have,
  but there are still some issues.
  Chen's method \cite{chen2018learning} is only applicable to raw images,
  so existing RGB color images cannot be enhanced.
  Yang's method \cite{yang2018image} generates intermediate high dynamic range (HDR)
  images from RGB images, and then high-quality LDR images are produced.
  However, generating HDR images from single images is an outstanding problem yet,
  as well-known \cite{eilertsen2017hdr, kinoshita2017fast}.
  For this reason, the performance of Yang's method is limited
  by the quality of intermediate HDR images.
  Furthermore, it is difficult to collect numerous training data with sufficient quality
  due to the narrow dynamic range of digital cameras.

  Thus, in this paper, we propose a novel image enhancement network trained by using HDR images.
  The proposed method can restore lost pixel values without generating intermediate HDR images.
  For training of a CNN used in the proposed method,
  we utilize existing HDR images.
  The use of HDR images makes it easy to collect training data
  for the CNN used in the proposed method.
  Specifically, both input images and target images are mapped
  from HDR images by tone mapping operations
  \cite{reinhard2002photographic, murofushi2013integer}.
  Because HDR images have much information that LDR images captured with cameras do not have,
  target LDR images mapped from HDR ones have better quality than
  that LDR ones directly captured with cameras.
  
  We evaluate the effectiveness of the proposed image enhancement network
  in terms of the quality of enhanced images in a number of simulations,
  where TMQI and NIQE are utilized as quality metrics.
  Experimental results show that the proposed method can restore lost pixel values.
  In addition, the proposed method outperforms
  state-of-the-art contrast enhancement methods
  in terms of both TMQI and NIQE.
  
\section{Related work}
  Here, we summarize conventional image enhancement methods
  including state-of-the-art ones,
  and recapitulate tone mapping methods used for the proposed method.
\subsection{Image enhancement}
  So far, a lot of image enhancement methods have been studied
  \cite{zuiderveld1994contrast, wu2017contrast,
  guo2017lime, fu2016weighted,
  kinoshita2017pseudo, kinoshita2018pseudo}.
  Among methods for enhancing images, HE has received
  the most attention because of its intuitive implementation quality and
  high efficiency.
  It aims to derive a mapping function such that the entropy of
  a distribution of output luminance values can be maximized,
  but HE often causes over-enhancement.
  To avoid the over-enhancement,
  numerous improved methods based on HE have also been developed
  \cite{zuiderveld1994contrast, wu2017contrast}.
  Another way for enhancing images is to use the Retinex theory \cite{land1977retinex}.
  Retinex-based methods \cite{guo2017lime, fu2016weighted} decompose images into
  reflectance and illumination, and then enhance images by manipulating illumination.
  However, these methods without CNNs
  cannot restore lost pixel values due to clipping and quantizing.
  
  Recently, a few CNN based methods were proposed \cite{chen2018learning, yang2018image}.
  Chen's method \cite{chen2018learning} provides high-quality RGB images
  from single raw images taken under low-light conditions,
  but this method cannot be applied to RGB color images.
  Yang's et al. \cite{yang2018image} proposed a method for enhancing RGB images
  by using two CNNs.
  This method generates intermediate high dynamic range (HDR)
  images from input RGB images, and then high-quality LDR images are produced.
  However, generating HDR images from single images is an outstanding problem yet,
  as well-known \cite{eilertsen2017hdr, kinoshita2017fast}.
  For this reason, the performance of Yang's method is limited
  by the quality of intermediate HDR images.
  In addition, collecting training data with sufficient quality is problematic
  due to the narrow dynamic range of digital cameras.

\subsection{Tone mapping}
  Tone mapping is an operation that generates an LDR image from an HDR image
  \cite{reinhard2002photographic, murofushi2013integer}.
  LDR images mapped from HDR images generally have higher quality
  than LDR ones directly captured with cameras.
  This is because HDR images have much information that
  LDR images captured with cameras do not have.
  
  Here, we summarize Reinhard's global operator which is one of typical tone mapping methods,
  as an example
  \cite{reinhard2002photographic}.
  Under the use of the operator, each pixel value in LDR image $I$
  is calculated from HDR image $E$ by
  \begin{align}
    I_{i, j} &= \hat{f}(X_{i, j}), \label{eq:map} \\
    \hat{f}(X_{i, j}) &= \frac{X_{i, j}}{1+X_{i, j}}, \label{eq:tmo}
  \end{align}
  where $(i, j)$ denotes a pixel and
  $X_{i, j}$ is given by using two parameters $a$ and $G(E)$ as
  \begin{equation}
    X_{i, j} = \frac{a}{G(E)} E_{i, j}. \label{eq:scale}
  \end{equation}
  In Reinhard's global operator, two parameters $a$ and $G(E)$ are used.
  $a \in [0, 1]$ determines brightness of an output LDR image $I$
  and $G(E)$ is the geometric mean of $E$.
  $G(E)$ is calculated as
  \begin{equation}
    G(E) = 
      \exp{
        \left(\frac{1}{|\myset{P}|}
          \sum_{(i, j) \in \myset{P}}
          \log{\left(\max{\left( E_{i, j}, \epsilon \right)}\right)}
        \right)
      },
    \label{eq:geoMeanEps}
  \end{equation}
  where $\myset{P}$ is the set of all pixels and
  $\epsilon$ is a small value to avoid singularities at $E_{i, j} = 0$.

  Tone mapping operations enables us not only to generate
  high-quality LDR images but also to simulate a digital camera, i.e., a virtual camera.
  For this reason,
  the proposed image enhancement network utilize
  LDR images mapped from HDR images as training data.

\section{Proposed image enhancement network}
\begin{figure}[!t]
    \centering
    \begin{subfigure}[t]{0.75\hsize}
      \centering
      \includegraphics[width=\columnwidth]{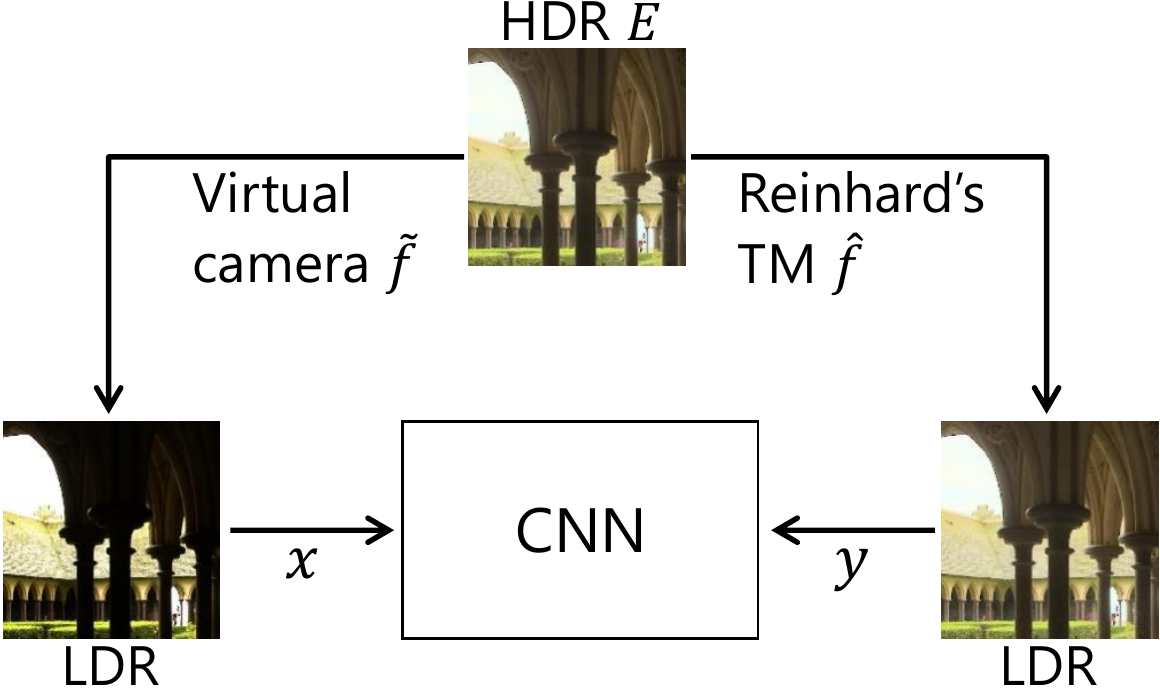}
      \caption{Training \label{fig:train}}
    \end{subfigure}\\
    \centering
    \begin{subfigure}[t]{0.95\hsize}
      \centering
      \includegraphics[width=\columnwidth]{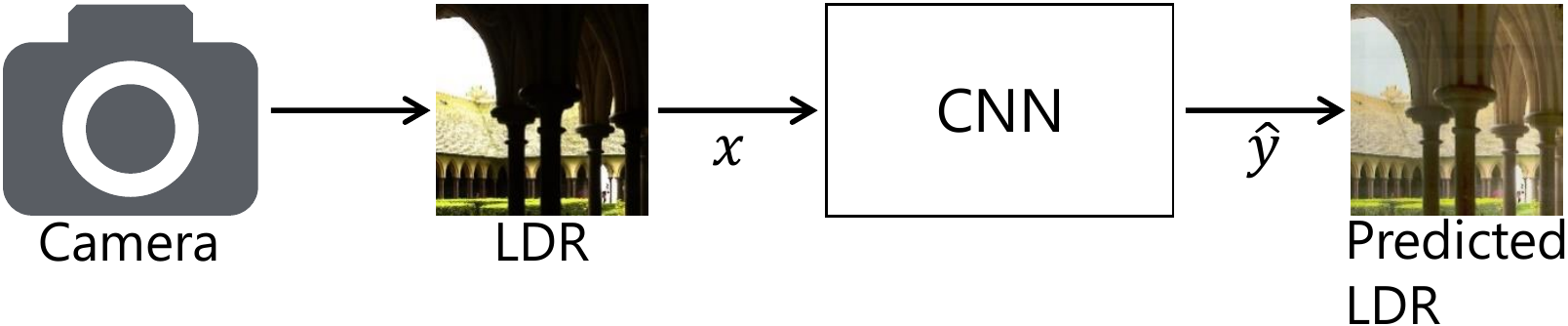}
      \caption{Predicting \label{fig:predict}}
    \end{subfigure}
    \caption{Proposed image enhancement method \label{fig:scheme}}
  \end{figure}

  Figure \ref{fig:scheme} shows an overview of
  our training procedure and predicting procedure.
  In the training, all input LDR images $x$ and target LDR images
  $y$ are generated from HDR images by
  a virtual camera \cite{eilertsen2017hdr} and Reinhard's global operator, respectively.
  This enables us not only to generate high-quality target images,
  but also to easily collect training data.
  In addition, the proposed method has a simpler network architecture
  than the conventional method \cite{yang2018image}
  because the proposed method do not generate intermediate HDR images.

  After the training, various LDR images are applied to the proposed network
  as input images,
  and then high-quality LDR images are generated by the network.
  Detailed training conditions will be shown in 3.2.

\subsection{Network architecture}
\begin{figure}[t]
    \centering
    \includegraphics[clip, width=0.95\hsize]{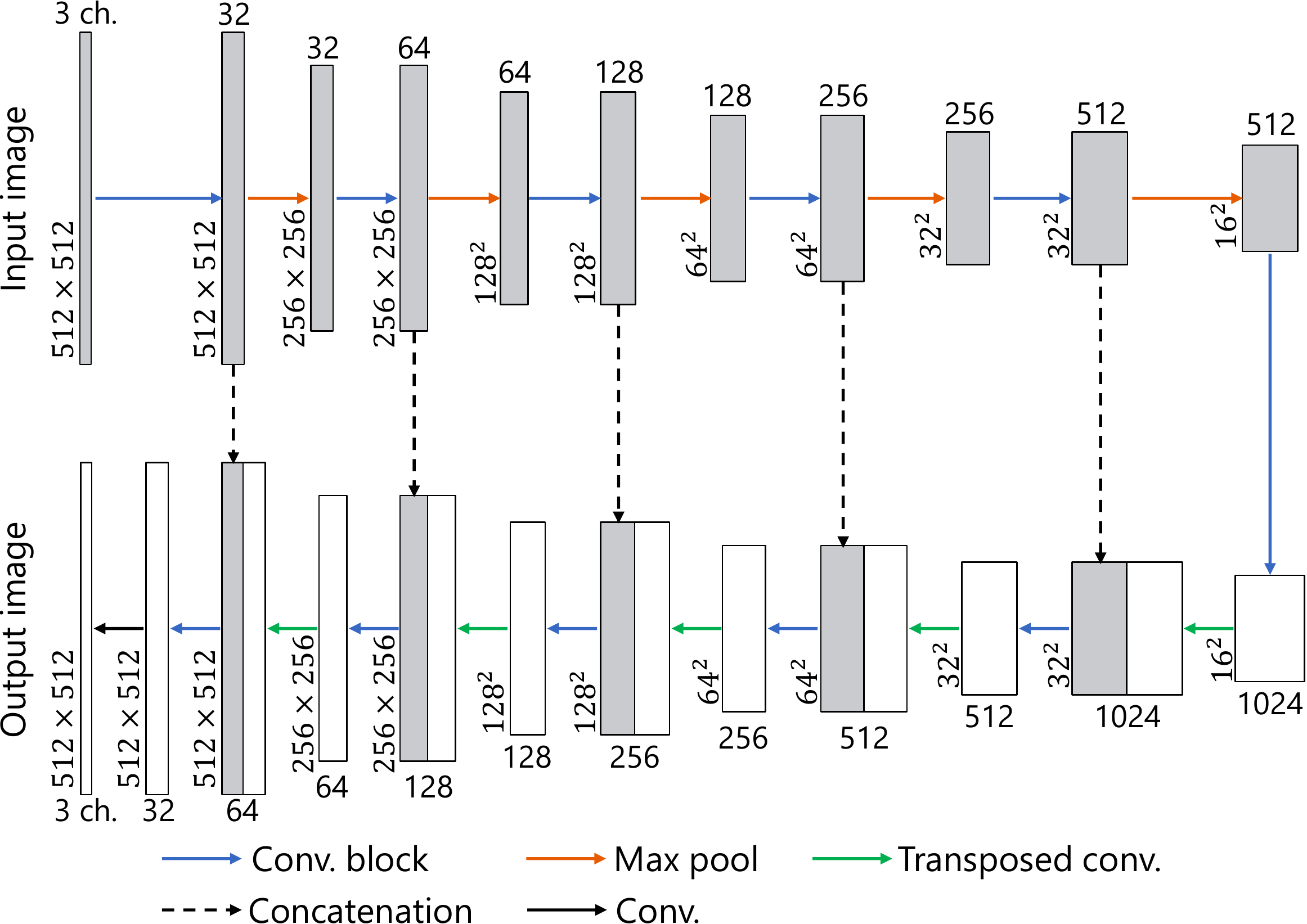}
    \caption{Network architecture.
      Each box denotes multi-channel feature map produced by each layer (or block).
      Number of channels is denoted on top or bottom of each box.
      Resolution of feature map is provided at left edge of box.
      \label{fig:network}}
  \end{figure}

  Figure \ref{fig:network} shows the network architecture of the CNN
  used in the proposed method.
  The CNN is designed on the basis of U-Net\cite{ronneberger2015unet}.
  The input of this CNN is a 24-bit color LDR image
  with a size of $512 \times 512$ pixels.

  Each convolutional block consists of two convolutional layers,
  in which the number of filters $K$ is commonly the same.
  From the first block to the last block,
  the numbers are given as $K = 32$, $64$, $128$, $256$, $512$, $1024$, $512$,
  $256$, $128$, $64$, and $32$,
  where all filters in convolutional blocks have a size of $3 \times 3$.
  Max pooling layers with a kernel size of $2 \times 2$ and
  a stride of $2$ are utilized for image downsampling.
  
  For image upsampling, transposed convolutional layers with a stride of $2$
  and a filter size of $4 \time 4$ are used in the proposed method.
  From the first transposed convolutional layer to the last one,
  the numbers of filters are $K = 512$, $256$, $128$, $64$,
  and $32$, respectively.
  Finally, an output LDR image is produced by
  a convolutional layer which has three filters with a size of $3 \times 3$.

  The rectified linear unit (ReLU) activation function \cite{glorot2011deep}
  is employed for all convolutional layers and transposed convolutional ones
  except the final convolutional layer.
  Further, batch normalization \cite{ioffe2015batch} is applied to
  outputs of ReLU functions after convolutional layers.
  The activation function of the final layer is a sigmoid function.

\subsection{Training}
  A lot of LDR images $x$ taken under various conditions
  and corresponding LDR ones $y$ with high-quality
  are needed for training the CNN in the proposed method.
  However, collecting these images with a sufficient amount is difficult.
  We therefore utilize HDR images $E$ to generate both input LDR images $x$
  and target LDR images $y$ by using a virtual camera \cite{eilertsen2017hdr}
  and Reinhard's global operator, respectively.
  For training, total 978 HDR images was collected from online available databases
  \cite{openexrimage, anyherehdrimage, hdrps, empa, nemoto2015visual, maxplanck}.
  
  A training procedure of our CNN is shown as follows.
  \begin{enumerate}[label=\roman*, align=parleft, leftmargin=*, nosep]
    \item Select eight HDR images from 978 HDR images at random.
    \item Generate total eight couples of input LDR image $x$ and target LDR one $y$
      from each of the eight HDR ones.
      Each couple is generated according to the following steps.
      \begin{enumerate}[label=(\alph*), align=parleft, leftmargin=*, nosep]
        \item Crop an HDR image $E$ to an image patch $\tilde{E}$ with $N \times N$ pixels.
          The size $N$ is given as a product of a uniform random number in range $[0.2, 0.6]$
          and the length of a short side of $E$.
          In addition, the position of the patch in the HDR image $E$
          is also determined at random.
        \item Resize $\tilde{E}$ to $512 \times 512$ pixels.
        \item Flip $\tilde{E}$ upside down with probability 0.5.
        \item Flip $\tilde{E}$ left and right with probability 0.5.
        \item Calculate exposure $X$ from $\tilde{E}$ by
          $X_{i, j} = \Delta t \cdot \tilde{E}$,
          where shutter speed $\Delta t$ is calculated as
          $\Delta t = 0.18 \cdot 2^v / G(\tilde{E})$ as in \cite{reinhard2002photographic}
          by using a uniform random number $v$ in range $[-4, 4]$.
        \item Generate an input LDR image $x$ from $X$
          by a virtual camera $\tilde{f}$, as
          \begin{equation}
            x = \tilde{f}(X) = \min \left((1+\eta)\frac{X^\gamma}{X^\gamma + \eta}, 1 \right),
          \end{equation}
          where $\eta$ and $\gamma$ are random numbers
          that follow normal distributions with mean $0.6$ and variance $0.1$
          and with mean $0.9$ and variance $0.1$, respectively.
        \item Generate a target LDR image $y$ from $\tilde{E}$
          by Reinhard's global operator (see eq. (\ref{eq:map}) to eq. (\ref{eq:geoMeanEps}))
          with parameter $a = 0.18$, where the parameter maps the average luminance of $y$ to
          the middle gray \cite{reinhard2002photographic}.
      \end{enumerate}
    \item Predict eight LDR images $\hat{y}$ from eight input LDR images $x$ by the CNN.
    \item Evaluate errors between predicted images $\hat{y}$ and target images $y$
      by using the mean squared error.
    \item Update filter weights $\omega$ and biases $b$ in the CNN by back-propagation.
  \end{enumerate}
  Note that steps \rnum{2}(f) and \rnum{2}(g) are applied to luminance of $\tilde{E}$,
  and then RGB pixel values of $x$ and $y$ are obtained
  so that ratios of RGB values of LDR images are equal to those of HDR images.

  In our experiments, the CNN was trained with 5000 epochs,
  where the above procedure was repeated 122 times in each epoch.
  In addition, each HDR image had only one chance to be selected,
  in step \rnum{1} in each epoch.
  He's method \cite{he2015delving}
  was used for initialization of the CNN
  In addition, Adam optimizer \cite{kingma2014adam} was utilized for optimization,
  where parameters in Adam were set as $\alpha=0.002, \beta_1=0.5$ and $\beta_2=0.999$.

\section{Simulation}
  We evaluated the effectiveness of the proposed method
  by using two objective quality metrics.

\subsection{Simulation conditions}
  In this experiment, test LDR images was generated
  from eight HDR images which were not used for training,
  according to steps from \rnum{2}(a) to \rnum{2}(f).

  The quality of LDR images $\hat{y}$ generated by the proposed method
  was evaluated by two objective quality metrics:
  the tone mapped image quality index (TMQI) \cite{yeganeh2013objective}
  and the naturalness image quality evaluator (NIQE) \cite{mittal2013making},
  where original HDR image $\tilde{E}$ was utilized as a reference for TMQI.

  The proposed method is compared with four conventional methods:
  histogram equalization (HE),
  contrast-accumulated histogram equalization (CACHE) \cite{wu2017contrast},
  simultaneous reflectance and illumination estimation (SRIE) \cite{fu2016weighted},
  and deep reciprocating HDR transformation
  \footnote{An approximate implementation
  at \url{https://github.com/ybsong00/DRHT} was utilized}
  (DRHT) \cite{yang2018image},
  where SRIE is a Retinex-based method and DRHT is a CNN-based one.
\subsection{Results}
\begin{figure*}[!t]
    \centering
    \begin{subfigure}[t]{0.16\hsize}
      \centering
      \includegraphics[width=\columnwidth]{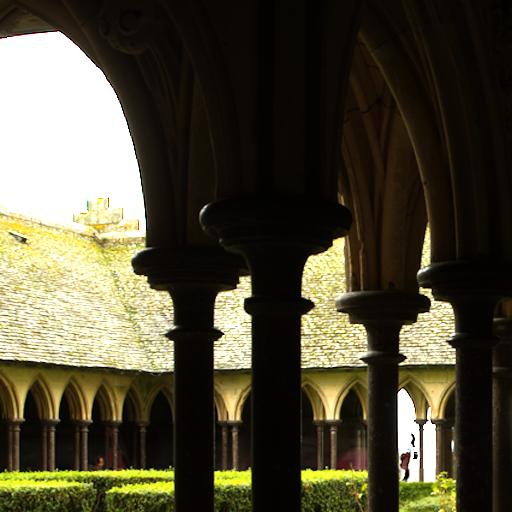}
      \caption{Input $x$\\
        TMQI: 0.8503\\
        NIQE: 4.975 \label{fig:input5}}
    \end{subfigure}
    \begin{subfigure}[t]{0.16\hsize}
      \centering
      \includegraphics[width=\columnwidth]{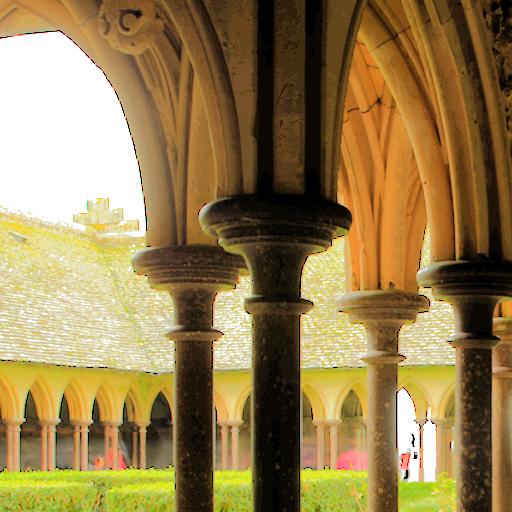}
      \caption{HE\\
        TMQI: 0.9352\\
        NIQE: 3.645 \label{fig:he5}}
    \end{subfigure}
    \begin{subfigure}[t]{0.16\hsize}
      \centering
      \includegraphics[width=\columnwidth]{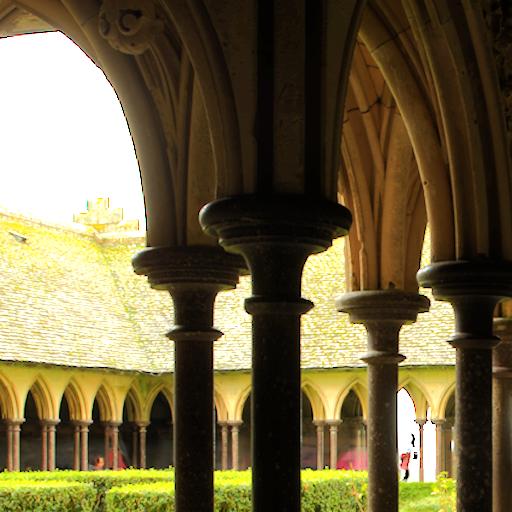}
      \caption{CACHE \cite{wu2017contrast}\\
        TMQI: 0.9425\\
        NIQE: 3.562 \label{fig:cache5}}
    \end{subfigure}
    \begin{subfigure}[t]{0.16\hsize}
      \centering
      \includegraphics[width=\columnwidth]{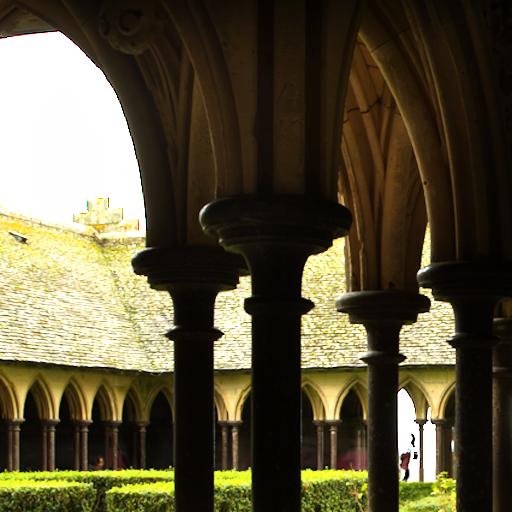}
      \caption{SRIE \cite{fu2016weighted}\\
        TMQI: 0.8823\\
        NIQE: 3.916 \label{fig:srie5}}
    \end{subfigure}
    \begin{subfigure}[t]{0.16\hsize}
      \centering
      \includegraphics[width=\columnwidth]{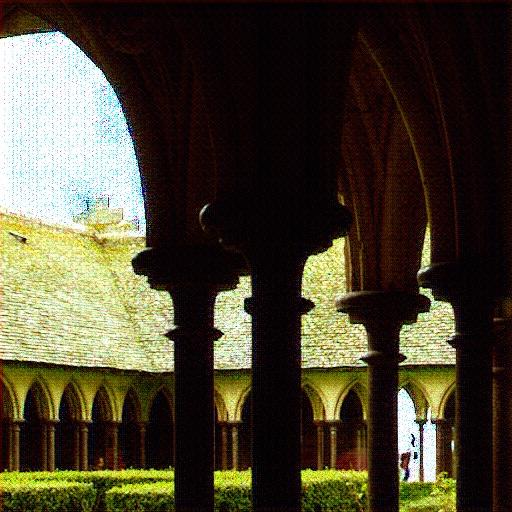}
      \caption{DRHT \cite{yang2018image}\\
        TMQI: 0.7827\\
        NIQE: 10.721 \label{fig:drht5}}
    \end{subfigure}
    \begin{subfigure}[t]{0.16\hsize}
      \centering
      \includegraphics[width=\columnwidth]{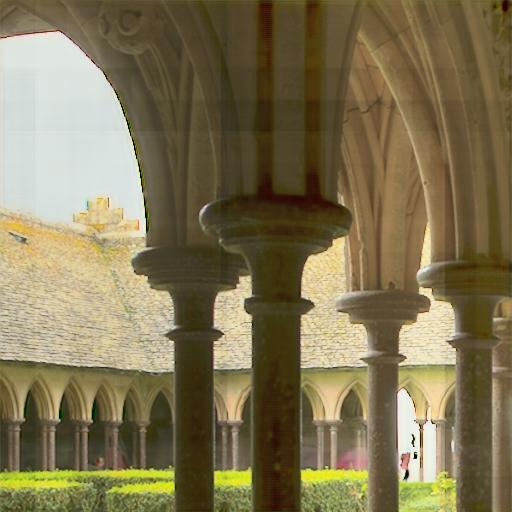}
      \caption{Proposed\\
        TMQI: 0.9463\\
        NIQE: 3.708 \label{fig:proposed5}}
    \end{subfigure}
    \caption{Experimental Results (Image 5) \label{fig:result5}}
  \end{figure*}

\begin{figure*}[!t]
    \centering
    \begin{subfigure}[t]{0.16\hsize}
      \centering
      \includegraphics[width=\columnwidth]{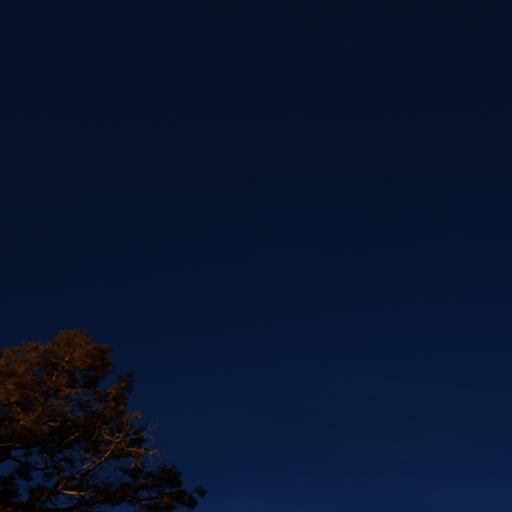}
      \caption{Input $x$\\
        TMQI: 0.4308\\
        NIQE: 5.462 \label{fig:input7}}
    \end{subfigure}
    \begin{subfigure}[t]{0.16\hsize}
      \centering
      \includegraphics[width=\columnwidth]{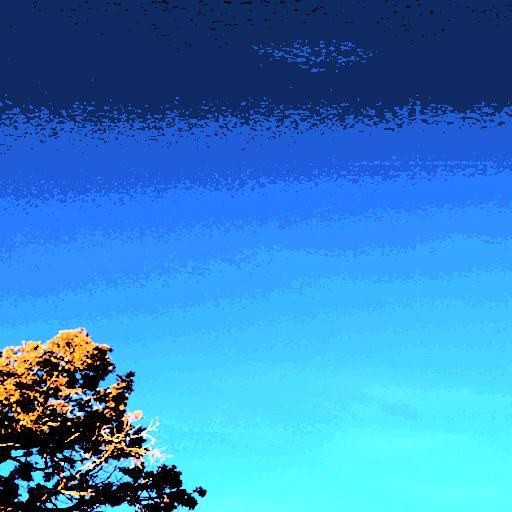}
      \caption{HE\\
        TMQI: 0.8705\\
        NIQE: 5.949 \label{fig:he7}}
    \end{subfigure}
    \begin{subfigure}[t]{0.16\hsize}
      \centering
      \includegraphics[width=\columnwidth]{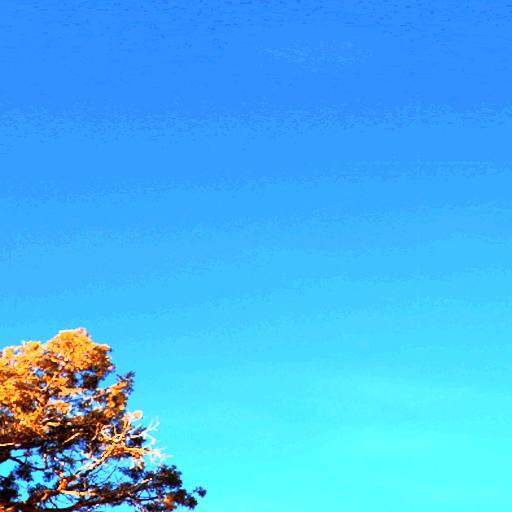}
      \caption{CACHE\cite{wu2017contrast}\\
        TMQI: 0.7537\\
        NIQE: 4.883 \label{fig:cache7}}
    \end{subfigure}
    \begin{subfigure}[t]{0.16\hsize}
      \centering
      \includegraphics[width=\columnwidth]{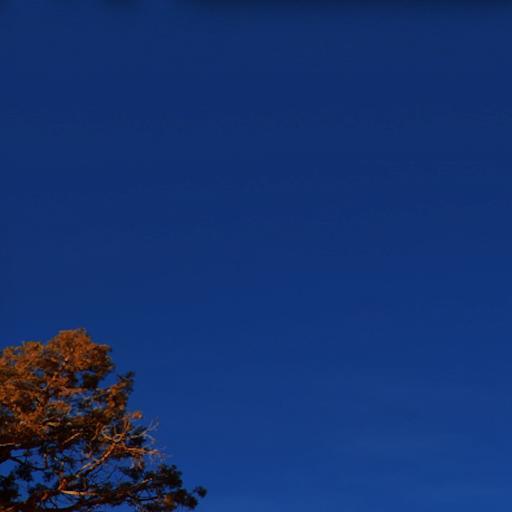}
      \caption{SRIE \cite{fu2016weighted}\\
        TMQI: 0.4830\\
        NIQE: 4.656 \label{fig:srie7}}
    \end{subfigure}
    \begin{subfigure}[t]{0.16\hsize}
      \centering
      \includegraphics[width=\columnwidth]{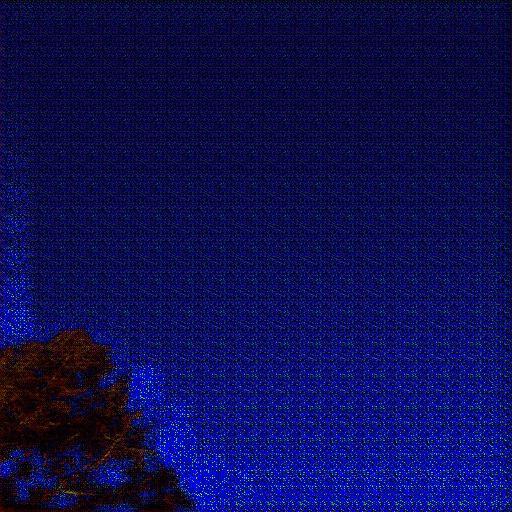}
      \caption{DRHT \cite{yang2018image}\\
        TMQI: 0.4027\\
        NIQE: 17.783 \label{fig:drht7}}
    \end{subfigure}
    \begin{subfigure}[t]{0.16\hsize}
      \centering
      \includegraphics[width=\columnwidth]{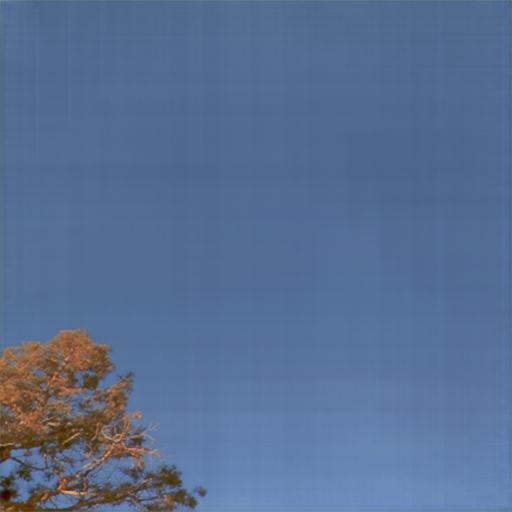}
      \caption{Proposed\\
        TMQI: 0.5491\\
        NIQE: 3.879 \label{fig:proposed7}}
    \end{subfigure}
    \caption{Experimental Results (Image 7) \label{fig:result7}}
  \end{figure*}

  Figures \ref{fig:result5} and \ref{fig:result7} show
  examples of HDR images generated by the five methods.

  From Fig. \ref{fig:result5},
  it is confirmed that the proposed method produced an higher-quality image
  which clearly represent dark areas in the image.
  The image generated by DRHT is still unclear
  due to the difficulty of generating an intermediate HDR image.
  Figure \ref{fig:result7} shows that
  the proposed method and SRIE produced high-quality images without banding artifacts.
  In contrast, images enhanced by HE and CACHE includes banding artifacts
  which are caused by quantized pixel values in the input image.
  This result denotes that the proposed method can restore lost pixel values
  due to the quantization.

  Tables \ref{tab:tmqi} and \ref{tab:niqe} illustrate results of objective assessment
  in terms of TMQI and NIQE.
  In case of TMQI, a larger value means a higher similarity between a target LDR image
  and an original HDR image.
  By contrast, a smaller value for NIQE indicates that a target LDR image has higher
  naturalness.
  As shown in Table \ref{tab:tmqi}, the proposed method provided the highest TMQI scores
  in the five methods for four images.
  HE also provided high TMQI scores,
  but HE cannot restore lost pixel values and it often causes the over-enhancement
  in bright areas.
  Moreover, as shown in Table \ref{tab:niqe},
  the proposed method also provided the best NIQE scores
  for six images (see Table \ref{tab:niqe}).
  For these reasons, the proposed method outperforms the conventional methods
  in terms of both TMQI and NIQE.
  \begin{table}[!t]
    \centering
    \caption{TMQI scores}
    {\small
    \begin{tabular}{l|c|c|c|c|c|c} \hline\hline
              & Input  & HE     & CACHE \cite{wu2017contrast}  & SRIE \cite{fu2016weighted}
              & DRHT \cite{yang2018image}  & Proposed \\ \hline
      Image 1 & 0.7858 & 0.8469          & 0.7427 & 0.7710 & 0.7682 & \textbf{0.9124} \\
      Image 2 & 0.6624 & 0.7422          & 0.6852 & 0.6624 & 0.6543 & \textbf{0.8870} \\
      Image 3 & 0.9394 & \textbf{0.9590} & 0.8580 & 0.8712 & 0.8809 & 0.8687          \\
      Image 4 & 0.6069 & \textbf{0.9215} & 0.9028 & 0.7441 & 0.5104 & 0.7693          \\
      Image 5 & 0.8503 & 0.9352          & 0.9425 & 0.8823 & 0.7827 & \textbf{0.9463} \\
      Image 6 & 0.8178 & 0.8502          & 0.9034 & 0.8808 & 0.7772 & \textbf{0.9104} \\
      Image 7 & 0.4308 & \textbf{0.8705} & 0.7537 & 0.4830 & 0.4027 & 0.5491          \\
      Image 8 & 0.7468 & \textbf{0.9031} & 0.8044 & 0.7461 & 0.6131 & 0.5969          \\
      \hline\hline
    \end{tabular}
    }
    \label{tab:tmqi}
  \end{table}
  \begin{table}[!t]
    \centering
    \caption{NIQE scores}
    {\small
    \begin{tabular}{l|c|c|c|c|c|c}\hline\hline
              & Input  & HE     & CACHE \cite{wu2017contrast}  & SRIE \cite{fu2016weighted}
              & DRHT \cite{yang2018image}  & Proposed \\ \hline
      Image 1 & 4.055 & 4.267 & \textbf{3.552} & 4.076 & 9.901  & 4.218          \\
      Image 2 & 4.740 & 5.064 & 4.919          & 4.723 & 12.145 & \textbf{2.870} \\
      Image 3 & 4.609 & 5.221 & 4.703          & 4.605 & 8.720  & \textbf{3.951} \\
      Image 4 & 6.754 & 6.918 & 4.904          & 4.191 & 17.873 & \textbf{4.127} \\
      Image 5 & 4.975 & 3.645 & \textbf{3.562} & 3.916 & 10.721 & 3.708          \\
      Image 6 & 5.118 & 4.556 & 4.698          & 4.920 & 10.523 & \textbf{4.220} \\
      Image 7 & 5.462 & 5.949 & 4.883          & 4.656 & 17.783 & \textbf{3.879} \\
      Image 8 & 5.803 & 5.869 & 4.289          & 5.775 & 13.515 & \textbf{4.112} \\
      \hline\hline
    \end{tabular}
    }
    \label{tab:niqe}
  \end{table}

  Those experimental results show that
  the proposed method is effective for enhancing single images.

\section{Conclusion}
  In this paper, a novel image enhancement network trained by using HDR images
  was proposed.
  The use of HDR images for training enables us to obtain
  higher-quality target images than that captured with cameras.
  Moreover, the proposed method has not only a higher performance
  but also a simpler network architecture, than the conventional CNN based method.
  Experimental results showed that the proposed method outperforms
  state-of-the-art conventional image enhancement methods in terms of TMQI and NIQE.
  In addition, visual comparison results demonstrated that
  the proposed method can restore lost pixel values, although conventional methods cannot.
  Because of space limitations, we have showed results for eight images,
  but the similar trend was confirmed for other images.

%


\end{document}